\documentclass[11pt]{article}
\usepackage{natbib}
\usepackage[lined,ruled,linesnumbered]{algorithm2e}
\usepackage{multirow}
\usepackage{color}
\usepackage{soul}
\usepackage{amsmath}
\usepackage{subfigure}

\usepackage[most]{tcolorbox}
\usepackage{multicol}
\usepackage{dcolumn}
\newcolumntype{d}[1]{D{.}{.}{#1}}

\begin{document}

\title{\bf When Hillclimbers Beat Genetic Algorithms in Multimodal Optimization}

\author{    {\bf Fernando G. Lobo}\\
            \small DEEI-FCT \& CENSE  -- Center for Environmental and Sustainability Research  \\
            \small  \& CHANGE -- Global Change and Sustainability Institute,\\ 
            \small Universidade do Algarve, Campus de Gambelas, 8005-139 Faro, Portugal\\
            \small flobo@ualg.pt
\and
           {\bf Mosab Bazargani}\\
            \small School of Electronic Engineering and Computer Science\\
            \small Queen Mary University of London, London E14FZ, UK\\
            \small m.bazargani@qmul.ac.uk
}

\maketitle

\let\oldnl\nl
\newcommand{\nonl}{\renewcommand{\nl}{\let\nl\oldnl}}

\hyphenation{op-tical net-works semi-conduc-tor}

\begin{abstract}

This paper investigates the performance of multistart next ascent hillclimbing and well-known evolutionary algorithms incorporating diversity preservation techniques on instances of the multimodal problem generator. This generator induces a class of problems in the bitstring domain which is interesting to study from a theoretical perspective in the context of multimodal optimization, as it is a generalization of the classical  \textsc{OneMax} and  \textsc{TwoMax} functions for an arbitrary number of peaks. 
An average-case runtime analysis for multistart next ascent hillclimbing is presented for uniformly distributed equal-height instances of this class of problems.
It is shown empirically that conventional niching and mating restriction techniques incorporated in an evolutionary algorithm are not sufficient to make them competitive with the hillclimbing strategy.

We conjecture the reason for this behaviour is the lack of structure in the space of local optima on instances of this problem class, which makes an optimization algorithm unable to exploit information from one optimum to infer where another optimum might be. When no such structure exist, it seems that the best strategy for discovering all optima is a brute-force one.

Overall, our study gives insights with respect to the adequacy of hillclimbers and evolutionary algorithms for multimodal optimization, depending on properties of the fitness landscape.
\end{abstract}


\section{Introduction}
\label{sec:intro}

It is commonly acknowledged that one of the advantages of evolutionary algorithms (EAs) with respect to classical optimization methods is their ability to escape local optima. By using a population of solutions, EAs can explore multiple basins of attraction simultaneously, and together with diversity preservation techniques, often referred in the literature as \emph{niching}, can maintain a diverse set of solutions and be able to locate multiple optima in a single run.

Research in niching methods has a long history in the field of evolutionary computation, starting with the work on \emph{crowding} introduced by~\cite{DeJong:75} and \emph{fitness sharing} by~\cite{Goldberg1987}, which set the basis for most of the subsequent niching methods proposed in the literature up to present times~\citep{Li2017,Sudholt2020}. The interest in niching methods is not restricted to the need of locating multiple optimal solutions for a problem.
Although often not explicitly mentioned, niching techniques were for a long time investigated with the goal of delaying convergence---and therefore maintaining multiple diverse solutions for a longer period of time---hoping that sub-parts of those solutions could be used by recombination operators to assemble better solutions, or to give time for an algorithm to discover gene linkage. Another key use of niching is in the context of multiobjective optimization where tradeoffs among multiple conflicting objectives are typically present and there is the need to preserve a diverse set of solutions~\citep{Horn:94,Deb:2002,Shir:09}, as well as in dynamic optimization problems where diversity is important to allow the algorithm to quickly react to changes in the environment~\citep{Branke:2001,Li:06,Blackwell:06}.

In the past decades there has been a growing interest in theoretical analysis of evolutionary algorithms, and in the more general class of randomized search algorithms.
Most of the work reported in the literature has been on simple artificial test functions.
These problems have the advantage of being easily understood, making them amenable to theoretical analysis, and also for being seen as bounding cases for algorithm performance on more complex problems. Take for example the \textsc{OneMax} problem which consists of maximizing the number of ones contained in a bitstring. This is a separable additive decomposable problem where each decision variable contributes equally to the overall fitness function, with no dependency among the decision variables. It is hard to imagine a simpler problem than \textsc{OneMax}, and that makes it an interesting bounding case for algorithm performance.

In the case of multimodal optimzation, a natural extension of \textsc{OneMax} is the \textsc{TwoMax} problem, also defined on bitstrings, with the fitness being given by either the number of ones or the number of zeros, whichever if greater. The problem has two optima located in the string of all zeros and in the string of all ones. Theoretical work on multimodal optimization with EAs has been using using the \textsc{TwoMax} problem, addressing the running time needed to locate the two optima~\citep{Friedrich2009,Oliveto:2019,Osuna2019,Osuna:2020,Osuna:2022}.

A natural extension to the \textsc{TwoMax} problem is to consider instances of the multimodal problem generator proposed by~\cite{DeJong:97} which is also defined over the bitstring domain. The generator constructs  problems with $m$ optima (peaks) whose locations are generated uniformly at random, and with the fitness of a string being determined by the hamming distance to its closest peak. Such an extension has been investigated by~\cite{Lobo2015} who suggested that an EA with niching is unlikely to perform better than a multistart hillclimber on this particular class of problems. Yet another extension is a related class of problems proposed by~\cite{Jansen2016} where the location of the peaks is also controllable, and which was empirically studied by~\cite{Osuna2018d}.

This paper is an extended and revised version of the work of~\cite{Lobo2015}, which was only published as a two-page poster at the 2015 Genetic and Evolutionary Computation Conference (GECCO). It looks at the performance of EAs and multistart hillclimbing when trying to obtain all optima of instances of the multimodal problem generator.
Herein we provide significant extensions upon the published GECCO poster. Concretely: (1) We provide a runtime analysis of the multistart next ascent hillclimbing, (2) the empirical study now includes the clearing technique embedded within a $(\mu+1)$ EA. This algorithm has been recently shown to be very effective due to its capacity to tunnel between local/global optima~\citep{Osuna2019,Osuna2018d}, (3) we present an idealized niching algorithm for this class of problems and argue that its performance should be close to the best that can be achieved by any practical niching algorithm on this class of problems, and (4) the empirical study is now conducted for various string lengths as opposed to a fixed one.

The paper is organized as follows. The next section describes the multimodal problem generator. Section~\ref{sec:msnahc} analyses the running time of multistart next ascent hillclimbing on this class of problems and confirms it with computer simulations. Section~\ref{sec:EAs} and~\ref{sec:experiments} assess if the incorporation of niching and mating restriction in evolutionary algorithms allows them to be competitive with the multistart hillclimbing strategy. Finally, Section~\ref{sec:conclusions} summarizes and presents the major conclusions of this work.

\section{The multimodal problem generator}
\label{sec:mpg}
The multimodal problem generator was proposed by~\cite{DeJong:97} and has been used by several researchers in subsequent studies. The generator creates problem instances with a certain number of peaks. For a problem with $m$ peaks, $m$ bitstrings of length $n$ are generated uniformly at random. Let us denote them by $peak_i$ with $i \in \{1,2,\ldots,m\}.$ Each of these strings is a peak (an optimum) in the landscape. Each $peak_i$ is assigned a height $h_i$. To evaluate a string $x$, we first locate its nearest peak $np(x)$ in Hamming space, with ties broken uniformly at random.

\begin{equation*}
\label{eq:near}
np(x) = \text{argmin}_{i=1..m} (D(x,peak_i))\;,
\end{equation*}

with $D(x,y)$ being the Hamming distance between strings $x$ and $y$, the number of positions where $x$ and $y$ have different bit values. Formally, $D(x,y) = \sum_{i=1}^{n} | x_i - y_i |$.

The fitness of a string $x$ is then given by the number of bits the string has in common with its nearest peak, divided by $n$, and scaled by the height of the nearest peak.

\begin{equation}
f(x) = \frac{n-D(x,peak_{np(x)})}{n}
 \cdot h(peak_{np(x)})\;.
\end{equation}


%

The difficulty of the problem depends on the number of peaks, their height distribution, and their location.

If the goal is to find all peaks then the problem becomes harder with increasing number of peaks. It also becomes harder if the peaks have disparate heights, as the low height peaks will be more difficult to detect and maintain, especially with evolutionary algorithms.
Finally, the location of the peaks in the search space can also make some problem instances more difficult than others. To see why this is so, consider a two-peak problem where the location of peaks are the strings 111$\ldots$1 and 000$\ldots$0. This is precisely the \textsc{TwoMax} problem discussed earlier. The peaks are maximally away from each other in Hamming space, and therefore, it is very difficult for a solution that is close to one of the peaks to move to a solution close to the other peak. The case when the optima are maximally away is an extreme situation. Another extreme situation is the case where the location of peaks are at a Hamming distance of 1 bit from each other. Under such a layout, the problem is trivial to solve because solutions can easily jump from peak to peak by a single bit flip.

If the location of peaks are generated uniformly at random, then the extreme (and close to extreme) situations are very unlikely to occur because the Hamming distance between peaks follows a Binomial distribution with $n$ Bernoulli independent trials, each with success probability $1/2$.
Either via the Normal approximation to the Binomial distribution or via the application of Chernoff bounds, it can be shown that the Hamming distance between any two peaks generated uniformly at random is strongly concentrated around its expected value $n/2$. See for example Lemma 10.8 in~\cite{Doerr2018k} which states that for all $\lambda \geq 0$,

$$
Pr\left[\left|D(x,y)-\frac{n}{2}\right| \geq \lambda\right] \leq 2 e^{(-2\lambda^2 / n)}.
$$

This property makes it difficult for randomized search algorithms, and in particular EAs, to exploit information from the search landscape to locate basins of attraction of undiscovered optima.

In the rest of the paper we address the computational effort to discover all optima for equal-height peak instances of the multimodal problem generator.

\section{Multistart next ascent hillclimbing and its analysis}
\label{sec:msnahc}

This section presents the application and analysis of a multistart next ascent hillclimbing algorithm (MS-NAHC) to solve instances of the multimodal problem generator with the target goal of discovering all optima. Starting from a solution generated uniformly at random, the algorithm climbs up the peak using next ascent hillclimbing (NAHC). Once there, it restarts from another solution generated uniformly at random, and keeps doing that until all optima are discovered.

NAHC systematically explores the neighbourhood of the current solution. It does so by generating a random permutation of the neighbours of the current solution, and visiting each neighbour in the order given by the random permutation. As soon as a neighbour $s$ is found with better fitness than the current solution, that neighbour $s$ becomes the current solution and a new random permutation is generated to visit the neighbours of the new current solution. This process is repeated until no neighbour improves upon the current solution. In this paper we consider the neighbourhood $\mathcal{N}$ of a solution $x$ to be the set of strings whose Hamming distance to $x$ is 1. Formally, $\mathcal{N}(x) = \{s : D(x,s) = 1 \}$.

Our NAHC implementation does not visit the neighbour of a string $x$ that was the previous current solution, because that solution can not possibly yield an improvement upon $x$. This way, NAHC only evaluates strings when strictly necessary. For completeness, we present pseudocode for NAHC and MS-NAHC as Algorithms~\ref{alg:nahc} and~\ref{alg:ms-nahc}, respectively.

\begin{algorithm}[!t]
	\SetKwInOut{Input}{Input}
	\SetKwInOut{Output}{Output}
	\DontPrintSemicolon

	\Output{A local optimum with respect to a given neighbourhood $\mathcal{N}$.}
	\BlankLine

	$x =$ solution generated u.a.r.\;
	$prev =$ NULL\;
	\Repeat{$hilltop$}{
		$hilltop =$ true\;
		\ForEach{$s \in \mathcal{N}(x) \backslash \{prev\}$}{       \label{alg:nahc:line:foreach}
			\If{$f(s) > f(x)$}{
				$hilltop =$ false\;
				$prev = x$\;
				$x = s$\;
				\textbf{break}\;
			}
		}
	}
	\Return $x$
	\caption{NAHC}
	\label{alg:nahc}
\end{algorithm}

\begin{algorithm}[!t]
	\SetKwInOut{Input}{Input}
	\SetKwInOut{Output}{Output}
	\DontPrintSemicolon

	\Output{A set of local/global optimal solutions.}
	\BlankLine

	$optima = \emptyset$\;
	\While{termination criterion not met}{
		$x =$ NAHC()\;
		\If{$x \not\in optima$}{
			add $x$ to $optima$\;
		}
	}
	\Return $optima$
	\caption{MS-NAHC}
	\label{alg:ms-nahc}
\end{algorithm}


\subsection{Average-case runtime analysis for solving equal-height uniformly distributed peaks with MS-NAHC}
\label{sec:analysis-MS-NAHC}

Let us analyze the average-case running time of MS-NAHC to find all optima of an $m$-peak equal-height instance with the location of peaks distributed uniformly at random. (Throughout the paper we always use $m$ to denote the number of peaks, and $n$ to denote the string length.)
The theoretical analysis makes the simplifying assumption that the peaks are equally likely to be reached on a single execution of NAHC. This simplifying assumption is used as an approximation to what happens in a real situation when the peaks are distributed uniformly at random, especially for large string length $n$ and few number of peaks. The analysis will be validated empirically in Section~\ref{sec:ms-nahc-experiments} and the results obtained confirm its accuracy. (See Figure~\ref{fig:MS-NAHC}.)

MS-NAHC calls NAHC repeatedly until all optima are found. In the process of doing so, it is natural that a given peak is discovered several times. Let us start by finding the expected number of times that NAHC needs to be called so that all $m$ optima are found, under the assumption of equal-sized basins of attraction.
The answer to this is given by the \emph{coupon collector's problem}~\citep{Motwani1995,Doerr2018k}.

In the coupon collector's problem there are $m$ coupon types and at each trial a coupon is chosen at random. Each coupon type is equally likely to be drawn. The expected number of trials needed to collect at least one copy of each type of coupon is $m \cdot H_m$, with $H_m = \sum_{i=1}^{m} 1/i$ being the $m$th Harmonic number which is asymptotically equal to $\ln m + O(1)$. A good approximation is given by $H_m \approx \ln m + \gamma$, where $\gamma \approx 0.5777216$ is the Euler-Mascheroni constant.

Under the assumption of equal-sized basins of attraction, each peak is equally likely to be reached on a single execution of NAHC. Therefore, each peak can be equated to a coupon type, and the coupon collector's problem is immediately applicable. Let $X_{\text{NAHC-EXECS}}$ be a random variable denoting the number of times NAHC is executed within the multistart strategy in order to discover the $m$ peaks. The expected value of
$X_{\text{NAHC-EXECS}}$ is given by,

\begin{align}
\label{eqn:restartsAllPeaks}
E[X_{\text{NAHC-EXECS}}] = m \cdot H_m = m  \sum_{i=1}^{m} 1/i = O( m \log m )
\end{align}

To find the overall running time, we just need to multiply $E[X_{\text{NAHC-EXECS}}]$ by the expected running time of NAHC to reach a peak starting from a solution generated uniformly at random. On a single peak problem, NAHC has the same running time as randomized local search (RLS), which is known to be $O(n \log n)$ for bitstrings of length $n$. The only difference between RLS and NAHC is in the way neighbours are explored: in RLS a neighbour of the current solution is sampled uniformly at random with replacement, while in NAHC neighbours are sampled without replacement. For completeness, we present the analysis for the case of NAHC.

Let $X_{\text{NAHC}}(n,d)$ be a random variable denoting the number of fitness evaluations needed by NAHC to find the optimum on a single peak problem instance assuming the starting string $x$ of length $n$ is at a Hamming distance $d$ from the peak. The expected value of $X_{\text{NAHC}}(n,d)$ is given by,

\begin{align}
\label{eqn:nahc}
E[X_{\text{NAHC}}(n,d)] = 1 + \frac{n}{d} + \sum_{i=1}^{d-1} \frac{n-1}{i}.
\end{align}

The first term in Equation~\ref{eqn:nahc} corresponds to the evaluation of the starting string. The second term, $n/d$, is the expected number of bit flips needed to get an improvement in fitness when the neighbourhood of $x$ is explored for the first time  (line~\ref{alg:nahc:line:foreach} of Algorithm~\ref{alg:nahc}).
The remaining terms are the expected number of bit flips needed for a fitness improvement when the current string $x$ is at distance 1, 2, $\ldots$, $d-1$, from the peak. These terms have $n-1$ instead of $n$ in the numerator because the previous current solution is not evaluated.

The distance $d$ is always less or equal to $n$. Therefore,

\begin{align}
\label{eqn:nahc2}
E[X_{\text{NAHC}}(n,d)] &\leq E[X_{\text{NAHC}}(n,d=n)] \nonumber \\
&= 2 + (n-1)  \sum_{i=1}^{n-1} 1/i \nonumber \\
& = 2 + (n-1)  ~  H_{n - 1}   \nonumber \\
& = O( n \log n ).
\end{align}

We are now ready to state the running time of MS-NAHC. Let $X_{\text{MS-NAHC}}(m,n)$ be a random variable denoting the number of fitness evaluations needed by MS-NAHC to find all peaks on an $m$ equal-height peak instance with bitstrings of length $n$.

\begin{align}
\label{eqn:ms-nahcAllPeaks}
E[X_{\text{MS-NAHC}}(m,n)] & \leq E[X_{\text{NAHC-EXECS}}] \cdot E[X_{\text{NAHC}}(n,d=n)] \nonumber \\
& = (m ~ H_m) \cdot (2 + (n-1)  ~  H_{n - 1})  \nonumber \\
& = O( m \log m ) \cdot O(n \log n ) \nonumber \\
& = O( m ~ n ~ \log m ~ \log n )
\end{align}

To get a good approximation for the expected number of fitness evaluations to locate all optima on this type of instances, it would be good to have an estimate of the expected distance $d$ of the starting string to its closest peak. For a single peak instance, $d=n/2$. For a larger number of peaks, however, the expected distance is less than $n/2$ because according to the definition of the multimodal problem generator, we first locate the nearest peak among the set of $m$ peaks.

Using $d=n/2$ as the expected distance of the starting string to its closest peak, should not result in a large error in terms of fitness evaluations because the number of iterations needed to get an improvement when the current string is far from the optimum is small. Nonetheless, a better estimate for the expected distance between the starting string and its closest peak can be obtained by considering first order statistics. The Hamming distance between bitstrings of length $n$ generated uniformly at random follows a Binomial distribution with $n$ Bernoulli independent trials, each with success probability $p=1/2$. For $n$ sufficiently large, the Binomial distribution can be approximated to the Normal distribution with mean $\mu = n p = n/2$ and standard deviation $\sigma = \sqrt{n p (1-p)} = \sqrt{n}/2$.

\cite{Royston:1982} quotes the following formula by~\cite{Blom:1958} as a good approximation to the expected value $E[Z_{r:m}]$ of the $r$-th order statistic of a set of $m$ samples of the Standard Normal distribution,

\begin{align}
\label{eqn:blom-orderStatisticApprox}
E[Z_{r:m}] \approx  \Phi^{-1} \Big(\frac{r - \alpha}{m - 2\alpha + 1}\Big)\;,
\end{align}

with $\alpha=0.375$ and $\Phi^{-1}$ being the inverse of the cumulative distribution function of the Standard Normal distribution. Using $r=1$ and $\alpha=0.375$ as suggested by Blom, we can obtain $E[Z_{1:m}]$, the expected value of the first order statistic (i.e., the minimum) of a set of $m$ samples of the Standard Normal distribution.

It is well known that any point $x$ from a Normal distribution with mean $\mu$ and standard deviation $\sigma$ can be converted to a point $z$ from the Standard Normal distribution by the formula $z = (x-\mu) / \sigma$. Conversely, any point $z$ from the Standard Normal distribution can be converted to a point $x$ from a Normal distribution with mean $\mu$ and standard deviation $\sigma$ by rearranging the previous formula: $x = \mu + \sigma z$~\citep{Hogg:1993}.
This means that for a $n$-bit instance, a good approximation for the expected distance $E[d(m,n)]$ between the starting string and the closest of $m$ peaks is given by $\mu + \sigma E[Z_{1:m}] = n/2 + \sqrt{n}/2 \cdot E[Z_{1:m}]$.

\subsection{Experimental verification}
\label{sec:ms-nahc-experiments}

We applied the MS-NAHC algorithm to equal-height instances of the multimodal problem generator with the location of the $m$ peaks generated uniformly at random. We generated 30 problem instances for each ($m$,$n$) combination, with $m \in \{20, 40, 80, 160, 320\}$ and $n \in \{50, 100, 200, 400\}$, and for each one, 100 independent runs were performed.

\begin{figure}
	\center
	\includegraphics[width=0.6\linewidth]{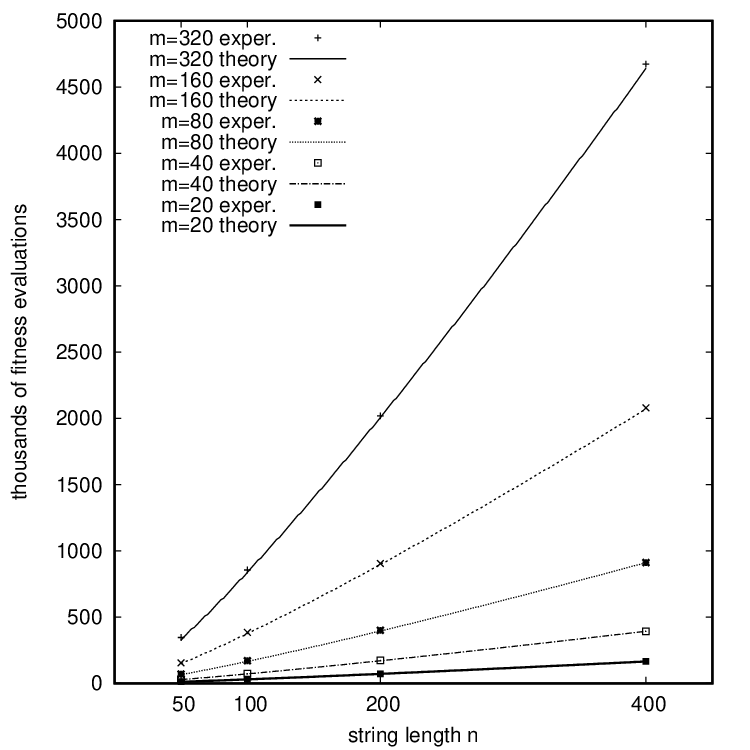}
	\caption{Average number of fitness evaluations needed by MS-NAHC to discover all optima on equal-heigh peak instances, along with the theoretical prediction using order statistics. The results are averaged over 30 problem instances generated uniformly at random for each $(m,n)$ combination, and using 100 independent runs for each one.}
	\label{fig:MS-NAHC}
\end{figure}

Figure~\ref{fig:MS-NAHC} shows the average number of fitness evaluations needed by MS-NAHC to discover all optima, along with the theoretical prediction using the first order statistic of a set of $m$ samples to obtain the expected distance of the starting string to its closest peak.
Table~\ref{tab:orderStat} shows the values of $E[Z_{1:m}]$ along with an approximation of the expected distance $E[d(m,n)]$ to the nearest of $m$ peaks on a $n$-bit problem instance, for $m \in \{20, 40, 80, 160, 320\}$ and $n \in \{50, 100, 200, 400\}$.

\begin{table}[h]


\centering
\caption{Approximation of the expected value of the first order statistic $Z_{1:m}$ of the Standard Normal distribution,
along with the expected distance $E[d(m,n)] \approx n/2 + \sqrt{n}/2 \cdot E[Z_{1:m}]$ to the nearest of $m$ peaks on a $n$-bit problem instance, for $m \in \{20, 40, 80, 160, 320\}$ and $n \in \{50, 100, 200, 400\}$.\label{tab:orderStat}}

\begin{tabular}{|c||d{3.2}|d{3.2}|d{3.2}|d{3.2}|d{3.2}|} \hline
                 & \multicolumn{1}{c|}{$m=20$} & \multicolumn{1}{c|}{$m=40$} & \multicolumn{1}{c|}{$m=80$} & \multicolumn{1}{c|}{$m=160$} & \multicolumn{1}{c|}{$m=320$} \\ \hline
$E[Z_{1:m}]$ & -1.87 & -2.16 & -2.42 & -2.66 & -2.89 \\ \hline
$E[d(m,50)]$ & 18.39 & 17.38 & 16.45 & 15.59 & 14.80 \\ \hline
$E[d(m,100)]$ & 40.66 & 39.22 & 37.91 & 36.70 & 35.57 \\ \hline
$E[d(m,200)]$ & 86.79 & 84.75 & 82.90 & 81.19 & 79.59 \\ \hline
$E[d(m,400)]$ & 181.32 & 178.44 & 175.81 & 173.39 & 171.14 \\ \hline
\end{tabular}


\end{table}

The theoretical prediction matches well with the experimental results, giving evidence that the simplifying assumption of considering equal-sized basins of attraction in the theoretical analysis is reasonable, and led to highly accurate results.

\section{Evolutionary algorithms with niching}
\label{sec:EAs}

An evolutionary algorithm which does not uses diversity preservation techniques is not able to find all optima on instances of the multimodal problem generator in a reasonable amount of time. Moreover, even if the goal is to reach the single global optimal solution on unequal-height peak instances, such an EA cannot reliably reach the best peak unless very large population sizes are used~\citep{Lobo:06a}. The cited work showed empirically that a multistart hillclimbing algorithm is more efficient than a so-called standard EA for solving instances of this class of problems. We shall now investigate if the incorporation of niching in an evolutionary algorithm can make them competitive with the multistart hillclimbing strategy on this class of problems. By maintaining diversity in a population of solutions the EA can maintain basins of attraction of several optima for long periods of time, hopefully allowing it to discover all optima in a single run. We implement and test the following four niched EAs.

\begin{itemize}
\item Restricted tournament selection.
\item Restricted tournament selection with mating restriction.
\item $(\mu+1)$ with deterministic crowding.
\item $(\mu+1)$ with clearing.
\end{itemize}

These algorithms have shown to be effective in practice and have been analyzed in theoretical and empirical studies (see~\cite{Li2017,Sudholt2020} for recent reviews.) However, as far as we know no theoretical analysis has been reported in the literature concerning the performance of these niched EAs on instances of the multimodal problem generator.

In addition to the above mentioned algorithms, we also implement an \emph{idealized niching} algorithm whose performance should be close to the best performance that any niching algorithm can achieve when solving instances of this class of problems. This algorithm, however, is unrealistic in practice because it knows where the niches are.


We shall now describe each algorithm in detail.

\subsection{Restricted tournament selection}
\label{sec:rts}

Restricted tournament selection (RTS)~\citep{Harik:95a} is a crowding-based niching algorithm that has been shown to be quite effective in practice. RTS incorporates the notion of local competition within a steady-state genetic algorithm (GA) forcing a new individual to compete with an existing population member that is similar to it. RTS works as follows. Two solutions are drawn uniformly at random from the population, call them $a$ and $b$. These two solutions undergo crossover and mutation giving rise to two new solutions, $a^{\prime}$ and $b^{\prime}$. Then, for each new solution ($a^{\prime}$ and $b^{\prime}$), scan $w$ individuals randomly chosen from the population and pick the one that is most similar to it. Call them $a^{\prime\prime}$ and $b^{\prime\prime}$. $a^{\prime}$ competes with $a^{\prime\prime}$, and the better one is allowed to stay in the population.
A similar competition takes place between $b^{\prime}$ and $b^{\prime\prime}$. $w$ is a parameter of RTS often referred to as the \emph{window size}. Pseudocode is shown as Algorithm~\ref{alg:rts}.

\begin{algorithm}[tbh]
  \SetKwFunction{MostSimilar}{most-similar}
  \SetKwFunction{Mutation}{mutation}
  \SetKwFunction{Crossover}{crossover}
  \DontPrintSemicolon
  $t = 0$\;
  Initialize $P_0$ with $\mu$ individuals chosen u.a.r.\;
  \While{termination criterion not met}{
    Choose $a \in P_t$ u.a.r.\;
    Choose $b \in P_t$ u.a.r.\;         \label{alg:rts:line:b}
    $a',b' =$ \Crossover{$a,b$}\;
    \ForEach{$s \in \{a',b'\}$}{
          $s' =$ \Mutation{$s$}\;
          $s'' =$ \MostSimilar{$P_t,w,s'$}\;
          \If{$f(s') > f(s'')$}{$P_{t} = P_{t} \setminus \{s''\}\cup \{s'\}$}
    }
    $P_{t+1} = P_{t}$\;
    $t = t+1$\;
  }
\caption{Restricted tournament selection (RTS)}
\label{alg:rts}
\end{algorithm}

\subsection{Restricted tournament selection with mating restriction}
\label{sec:rtsmr}

RTS does not have a mating restriction mechanism preventing solutions from different basins of attraction to mate with each other. On instances of the multimodal problem generator, crossing solutions near the top of distinct peaks is likely to produce so-called \emph{lethal} solutions which are far from any peak. As observed by~\cite{Lobo:06a}, crossover is only beneficial in these problem instances when it crosses solutions belonging to the same basin of attraction. To address this issue we implement a mating restriction mechanism on top of RTS. We name the resulting method as RTSMR. As opposed to RTS, only one solution is randomly chosen from the population, call it $a$. Ideally $a$ should mate with a solution that is not too far away from it, i.e., a solution in the same basin of attraction. The obvious way to achieve this is to implement the same method employed by RTS for finding a similar individual to compete with, and use it for the mating phase as well. As such, instead of picking the second solution $b$ uniformly at random, we scan $w$ individuals at random from the population and pick the one that is most similar (but whose distance to it is at least 2 bits) to mate with it. The 2-bit minimum distance restriction is used because crossing two bitstrings whose Hamming distance is less than 2 always produces children identical to the parents, regardless of the crossover operator used. The remaining part of the algorithm is exactly the same as in RTS. In other words, RTSMR implements both a mating restriction policy as well as a competition/replacement restriction. Pseudocode is shown as Algorithm~\ref{alg:rtsmr}. Both RTS and RTSMR call auxiliary functions shown as Pseudocode Listing~\ref{alg:most-similar}.

\begin{algorithm}[tbh]
  \SetKwFunction{MostSimilar}{most-similar}
  \SetKwFunction{MostSimilarTwo}{most-similar2}
  \SetKwFunction{Mutation}{mutation}
  \SetKwFunction{Crossover}{crossover}
  \DontPrintSemicolon

   \tcc{Pseudocode is the same as RTS with line \ref{alg:rts:line:b} in Algorithm~\ref{alg:rts} replaced by the line below.}
    \nonl $b =$ \MostSimilarTwo{$P_t,w,a$}\; \nonl
\caption{Restricted tournament selection with mating restriction (RTSMR)}
\label{alg:rtsmr}
\end{algorithm}

\begin{algorithm}[tbh]
  \SetAlgorithmName{Pseudocode Listing}{}{}
  \SetKwFunction{MostSimilar}{most-similar}
  \SetKwProg{Fn}{Function}{}{}
  \DontPrintSemicolon
  \Fn{\MostSimilar{$P$, $w$, $x$}}{
       $Q = $ Choose  $w$ individuals $ \in P$ u.a.r. without replacement\;
        \KwRet $\text{argmin}_{s \in Q} D(s,x)$\;
  }
\;
  \SetKwFunction{MostSimilarTwo}{most-similar2}
  \SetKwProg{Fn}{Function}{}{}
  \DontPrintSemicolon
  \Fn{\MostSimilarTwo{$P$, $w$, $x$}}{
       $Q = $ Choose  $w$ individuals $ \in P$ u.a.r. without replacement\;
       $T =  \{ s \in Q : D(x,s) > 1 \}$\;
       \eIf{$T \neq \emptyset$}
            { \KwRet $\text{argmin}_{s \in T} D(s,x)$}
            { \KwRet $s \in Q$ u.a.r.}
  }
\caption{Auxiliary functions called by RTS and RTSMR}
\label{alg:most-similar}
\end{algorithm}

\subsection{$(\mu+1)$ with deterministic crowding}
\label{sec:mu1DC}

This algorithm uses mutation alone and has a replacement strategy that enforces a crowding-like mechanism when mutation rates are low. The algorithm is very simple. A solution $a$ is drawn uniformly at random from the population. That solution undergoes mutation yielding a new solution $a^{\prime}$. Then $a^{\prime}$ competes with $a$ and whichever is best is allowed to stay in the population. With a low mutation rate, $a$ and $a^{\prime}$ should be similar to each other, and the competition between them enforces a crowding mechanism, just like in RTS. This algorithm is referred to as $(\mu+1)$~EA with deterministic crowding, and it has also been explored by~\cite{Friedrich2009}. We shall henceforth use the acronym $(\mu+1)$-DC to refer to it. Pseudocode is shown as Algorithm~\ref{alg:mu1DC}.

\begin{algorithm}[tbh]
  \SetKwFunction{Mutation}{mutation}
  \DontPrintSemicolon
  $t = 0$\;
  Initialize $P_0$ with $\mu$ individuals chosen u.a.r.\;
  \While{termination criterion not met}{
    Choose $x \in P_t$ u.a.r.\;
     $y =$ \Mutation{$x$}\;
  \eIf{$f(y) \ge f(x)$}{$P_{t+1} = P_{t} \setminus \{x\}\cup \{y\}$}{$P_{t+1} = P_{t}$}
	$t = t+1$
}
\caption{($\mu$+1) with deterministic crowding}
\label{alg:mu1DC}
\end{algorithm}

\subsection{ $(\mu+1)$ with clearing}
\label{sec:mu1CL}

Clearing is a niching method inspired by the classical fitness sharing idea where the real fitness of an individual was derated based on the amount of similarity of that individual with other population members. The clearing method~\citep{Petrowski1996}, however, takes a more aggressive approach and gives all the resources within a niche to the best individual in that niche, the so-called \emph{winner}. The remaining individuals of that niche have their fitness reset to a value lower than the lowest fitness value in the search space, and are said to be \emph{cleared} by the winner individual.

Similarly to fitness sharing, a distance threshold parameter $\sigma$ needs to be specified and is used to determine if individuals belong to the same niche. The method can be generalized to support up to $\kappa$ winners per niche, with $\kappa$ being a parameter referred to as the \emph{niche capacity}.

\cite{Osuna2019} embedded the clearing method within a $(\mu+1)$ EA. We use the acronym $(\mu+1)$-CL to refer to it. The pseudocode is shown in Algorithm~\ref{alg:mu1CL} which uses Algorithm~\ref{alg:clearing-procedure} for the clearing procedure.
(We use $-\infty$ in the pseudocode to denote a value lower than the lowest fitness value in the search space.)


\begin{algorithm}[tb]
\SetKwInOut{Input}{input}\SetKwInOut{Output}{output}
\SetKwInOut{Note}{note}
\Input{A population $P$}
\Output{Applies the clearing procedure to $P$.}
\Note{$-\infty$ denotes a constant value lower than the lowest fitness value in the search space.}
  Sort $P$ in decreasing order of fitness.\\
  \For{$i = 1$ \textbf{\upshape{}to} $|P|$}{
    \If{$f(P[i]) > -\infty$}{
        $\mathrm{winners} = 1$.\\
        \For{$j = i + 1$ \textbf{\upshape{}to} $|P|$}{
          \If{$f(P[j]) > -\infty$ \textbf{\upshape{}and} $D(P[i], P[j]) < \sigma$}{
            \eIf{$\mathrm{winners} < \kappa$}{$\mathrm{winners} = \mathrm{winners} + 1$}{$f(P[j]) =  -\infty$}
          }
      }
    }
  }
\caption{Clearing procedure}
\label{alg:clearing-procedure}
\end{algorithm}


\begin{algorithm}[tbh]
  \SetKwFunction{Mutation}{mutation}
  \DontPrintSemicolon
  $t = 0$\;
  Initialize $P_0$ with $\mu$ individuals chosen u.a.r.\;
  \While{termination criterion not met}{
    Choose $x \in P_t$ u.a.r.\\
    $y =$ \Mutation{$x$}\;
    Apply the clearing procedure to $P_t \cup \{y\}$  $\quad$ \tcp{Algorithm~\ref{alg:clearing-procedure}}
    Choose $z \in P_t$ with worst fitness after clearing u.a.r.\;    
  \eIf{$f(y) \ge f(z)$\label{ds:li:eval}}{$P_{t+1} = P_{t} \setminus \{z\} \cup \{y\}$}{$P_{t+1} =  P_{t}$}
	$t = t+1$
}
\caption{($\mu$+1) with clearing}
\label{alg:mu1CL}
\end{algorithm}

This procedure processes the individuals in the population in decreasing order of fitness. For each individual, if it has not been cleared, it is declared a winner. Then the procedure iterates through all remaining individuals (i.e., those with lower or equal fitness) that haven't been cleared yet and that are within a distance of $\sigma$, adding them to its niche until $\kappa$ winners have been found, and clearing the remaining ones.

The clearing procedure has a time complexity of $\Theta(\mu^2)$, and is called within the main loop of $(\mu+1)$-CL for every solution that is generated. This makes it prohibitive to perform experiments with very large population sizes. To address this issue we designed a more efficient implementation of the $(\mu+1)$-CL algorithm when the niche capacity $\kappa=1$, decreasing the running time from $\Theta(\mu^2)$ to $\Theta(\mu)$ per generated individual. The idea is to maintain the population $P$ sorted in descending order of fitness from iteration to iteration. When a new individual $y$ is generated, we need to apply the clearing procedure to $P~\cup ~y$. We do that without explicitly inserting $y$ in $P$, we just need to find the location of $y$ in $P$ in sorted order. (We can do so in logarithmic time with binary search.) Let $yp$ be the desired location. That is, $f(P[i]) > f(y) ~ \forall i \in [0..yp-1]$ and $f(P[i]) \leq f(y) ~ \forall i \in [yp..\mu-1]$. Then, $y$ can only be cleared by someone fitter than $y$ (i.e., some string in $P[0..yp-1]$), and $y$ can only clear strings with lower fitness (i.e., strings in $P[yp..\mu-1]$). This can be done in $\Theta(\mu)$ time. The location of the cleared individuals can be recorded so that the population can be rearranged and maintained in sorted order for the next iteration, and that can also be done in linear time.

In this faster version, the regular clearing procedure (Algorithm~{\ref{alg:clearing-procedure}) is only executed once, immediately after the initialization of the population. Appendix~\ref{sec:eClearing} shows pseudocode as Algorithm~\ref{alg:Efficient-mu1CL} which calls Algorithm~\ref{alg:efficient-clearing-procedure}.\footnote{Our concrete implementation of the pseudocode uses further optimizations, but they don't change the $\Theta(\mu)$ running time per generated individual, and they are omitted from the pseudocode  for the sake of simplicity.}

\subsection{Idealized niching algorithm}
\label{sec:iNiching}


We shall now present an idealized niching algorithm whose performance should be close to the best that any niching algorithm can achieve on instances of the multimodal problem generator.

When solving an $m$-peak instance, suppose that at the end of each generation the idealized niching algorithm is able to cluster the population into sub-populations, each containing the solutions that are at the basin of attraction of a particular peak. In other words, the idealized algorithm knows where the niches are. (When evaluating a solution we have to compute the nearest peak, and therefore obtain the desired information.) This is of course an unrealistic assumption to hold in practice for an arbitrary problem, but that's precisely the reason why we call it an \emph{idealized} algorithm. Under this assumption, the idealized algorithm could restrict its EA operations (selection, crossover, mutation, replacement) to occur only within members of each sub-population, in effect being equivalent to having $m$ separate EAs. We use the acronym iNiching to refer to this algorithm. Pseudocode is shown as Algorithm~\ref{alg:iNiching}.

\begin{algorithm}[tbh]
  \SetKwFunction{Generation}{Generation}
  \SetKwFunction{np}{np}
  \DontPrintSemicolon
  $t = 0$\;
  Initialize $P_0$ with $\mu$ individuals chosen u.a.r.\;
  \While{termination criterion not met}{
    \tcp{Partition $P_t$ in sub-populations}
    create empty sub-populations $Q_i$, with $i$ in [1..$m$]\;
    \ForEach{$x \in P_t$}{
         insert $x$ in $Q_{\np(x)}$. $\quad$ \tcp{$\np$ is the nearest peak function}
    }
    \tcp{Run an EA generation in each $Q_i$}
    \ForEach{$Q_i$}{
         $R_i = \Generation(Q_i)$
    }
    $P_{t+1} = \bigcup_{i=1}^{m} R_{i}$\;
    $t = t+1$\;
  }
\caption{Idealized niching algorithm for instances of the multimodal problem generator}
\label{alg:iNiching}
\end{algorithm}


\section{Experiments}
\label{sec:experiments}

This section presents the experimental setup and the results obtained from the application of the five niched EAs on the first set of instances reported in Section~\ref{sec:ms-nahc-experiments} for MS-NAHC, for each ($m$,$n$) combination. 

On all runs we imposed a maximum number of fitness evaluations, upon which we considered a run to be unsuccessful. This limit was set to the three times the number of fitness evaluations needed by the worst of the 100 independent runs of MS-NAHC when solving the instance with the same size $n$ with the largest number of peaks ($m=320$), rounded up to the closest million. This resulted in using the following limits: 2, 5, 9, and 19 million fitness evaluations, for problem lengths 50, 100, 200, 400, respectively.

We did our best to use near-optimal parameter settings for the EAs so that they could perform as best as possible. We discuss the setup below:

\subsection{Setup}
\label{sec:setup}

\subsubsection{Population size}


We use the bisection method first proposed by~\cite{Sastry:01a} to find a population size that is able to reach the target goal (find all optima) on 100/100 independent runs, while spending the least number of fitness evaluations possible.

The bisection method can be used in controlled experiments when the target goal is known in advance, which is the case here. We assume the number of fitness evaluations taken by an EA to reach all optima, is a unimodal function with respect to population size. This justifies the use of the bisection method.

Our implementation of the method works as follows. Starting from a very small population size, it keeps doubling it until the algorithm is able to reach all optima in 100/100 independent runs. Once such a large enough population size is found, the number of evaluations taken is recorded. We then continue to double the population size as long as the corresponding number of fitness evaluations needed to reach all optima decreases. Let the resulting population size be $\mu$. Then we know that the ideal population size is between $low = \mu/2$ and $high = \mu$. The bisection method then tries a population size in the middle of the two, $mid = (high+ low)/2$, and refines the $low$ or $high$ bounds accordingly depending on the outcome of the sequence of independent runs performed with the new size $mid$. If those runs fulfill the desired quality criterion with fewer fitness evaluations, $high$ becomes $mid$, otherwise $low$ becomes $mid$. This process is repeated until $high$ and $low$ are within a certain threshold factor of each other. (We use a 0.1 factor in our experiments.) At that point the value $high$ is returned as the method's guess for the best  population size that ensures the target goal. The sequence of runs performed with that population size are then used for measuring the performance of the algorithm. If a bisection run is not able to find a population size that can discover all optima in 100/100 independent runs, it is considered unsuccessful.

We performed 30 independent executions of the bisection method, per algorithm and per problem instance, and report the results over the successful ones. That is, if there are $ns$ successful bisection runs for a given algorithm on a given problem instance, then a total of $100 \times ns$
fitness evaluation values are used to measure the performance.

\subsubsection{Crossover and mutation}

RTS, RTSMR, and iNiching are the only ones that use crossover. We use uniform crossover on the experiments~\citep{Syswerda:89}. On instances of the multimodal problem generator, crossover is only effective when crossing strings in the same basin of attraction~\citep{Lobo:06a}. In such cases it is as if the problem had only a single peak and that would make it equivalent to the classical \textsc{OneMax} problem for which uniform crossover provides better mixing and faster convergence than $k$-point crossover~\citep{Thierens:99}. For iNiching we use $P_c = 0.8$. For RTS and RTSMR we test three crossover rates $P_c \in \{0.0, 0.5, 0.8\}$.

For all EAs the mutation operator is independent bit-flip with probability $1/n$.

Our implementation of the algorithms was done in such a way that new individuals were only evaluated if that was absolutely required. Whenever a newly created individual is identical to one of the parent individuals, no fitness evaluation is spent. Similarly, during the local competition on RTS and RTSMR, if the two competing individuals are identical, no fitness evaluation is spent.

\subsubsection{Other algorithm specific configuration}

For RTS and RTSMR, the window size $w$ was set to 4 times the number of peaks, or to the population size (whichever was minimum), following the recommendations given by~\cite{Harik:95a}.

For the $(\mu+1)$-CL we use a niche capacity of $\kappa=1$, which corresponds to maximum clearing. The clearing radius $\sigma$ was set to ensure that no two peaks clear each other, with high probability. To do so, we set $\sigma$ to be five standard deviations away from the expected distance between two bitstrings of length $n$ generated uniformly at random: $\sigma = \lfloor n/2 - 5 \sqrt{n}/2 \rfloor$.

For iNiching, the EA generation that is run in each sub-population consists of binary tournament selection without replacement, followed by crossover, mutation, and 50\% worst replacement. An exception occurs if the sub-population contains a single individual: in that case an iteration of a $(1+1)$~EA is performed.

\subsubsection{Additional stopping criterion}

The bisection method requires the execution of a large number of runs with potentially very large population sizes, and that has a significant impact on the time needed to complete the experiments. To speed up the experiments we make use of theoretical insights to halt the execution of a run under certain conditions, which we explain next.

To facilitate the exposition, we say that a population $P$ covers the basin of attraction of $peak_i$ if it contains at least one solution whose nearest peak is $peak_i$. Formally, $P$ covers $peak_i$ if and only if $np(P[x]) = i$ for some $x \in P$.

With the exception of $(\mu+1)$-CL, all the other EAs stop the execution of a run when the population does not cover the basin of attraction of all peaks. The reason for doing this is because once evolution starts, selection favors fit solutions which are the ones nearest to the top of the various peaks, and those are the solutions that will be present in the population at any given time. Suppose the population does not cover the basin of attraction of $peak_i$. Then, the probability of obtaining a solution at the basin of attraction of such a peak, and having that solution survive in the population, becomes arbitrarily small: many bits would need to be flipped at once so that the mutation operator could yield such a solution, and a very large time would be required to do so. Likewise, crossing two solution from different basins of attraction could yield an offspring at the basin of attraction of $peak_i$ but such a solution will be with high probability at the bottom of the hill and will not survive selection. In summary, once a basin of attraction is lost we can safely halt the run and consider it to be unsuccessful.
(This observation is confirmed in preliminary experiments that we did where we allowed the run to proceed once a basin of attraction was lost from the population: no solution at that basin of attraction ever emerged during the remaining maximum number of evaluations allowed.)

The $(\mu+1)$-CL algorithm is an exception because it is able to discover new basins of attraction even when they are not present in the population. As explained by~\cite{Osuna2019}, clearing assigns fitness to the  \emph{winners} of each niche, and conducts a random walk among the \emph{cleared} individuals. That random walk eventually discovers an individual at a new basin of attraction, and that individual will survive because it will be a winner of a new niche~\citep{Osuna2019}.

We make use, however, of a theoretical result obtained by~\cite{Witt2006} on the analysis of the running time of the $(\mu+1)$ EA on \textsc{OneMax} to estimate a minimum number of fitness evaluations needed by $(\mu+1)$-CL to discover a peak whose basin of attraction has not yet been covered. To discover such a peak, the algorithm first has to discover a solution at the basin of attraction of that peak, and then it has to climb to the top. This implies that a lower bound for the running time of $(\mu+1)$ EA on \textsc{OneMax} is also a lower bound for the $(\mu+1)$-CL to reach a peak whose basin of attraction has not yet been covered. An asymptotic lower bound of $\Omega(\mu n)$ is known to hold when $\mu$ is larger than $n$~\citep{Witt2006}. Although no formal proof has been published, Witt described in a personal communication the intuition behind a proof sketch using the family tree technique~\citep{Witt2006} arguing that a lower bound of $(1/(2e)-\epsilon)\mu n$ fitness evaluations hold with high probability, for an arbitrarily small constant $\epsilon>0$~\citep{Witt2020}.

We make use of this bound to halt a $(\mu+1)$-CL run when appropriate. Concretely, if the population does not cover a basin of attraction, and the remaining number of allowed fitness evaluations is less than $1/(2e) \mu n$, we halt the algorithm and declare the run to be unsuccessful.

\subsection{Results}
\label{sec:results}


Figures~\ref{fig:equal-height-m020} through \ref{fig:equal-height-m320} show the results obtained for increasing number of peaks $m \in \{20, 40, 80, 160, 320\}$, respectively. For each $m$ value, four box plots are shown corresponding to string length $n \in \{50, 100, 200, 400\}$. Each box plot presents the results of all EAs along with the results of MS-NAHC. A missing algorithm in a box plot means that on all 30 bisection runs, the algorithm failed to discover all optima under the allowed maximum number of fitness evaluations.

The results show that MS-NAHC is faster than all EAs with the exception of iNiching. This is confirmed by Mann-Whitney statistical significance tests conducted for all $(m,n)$ combinations, all of them giving a $p$-value $=0$ and $U=0$ statistic: for all $(m,n)$ combinations, the number of fitness evaluations taken by the worst run of MS-NAHC is less than the number of fitness evaluations taken by the best run of the competing EA.

iNiching is the only algorithm that is competitive with MS-NAHC. This is not surprising because iNiching knows where the niches are and takes explicit advantage of that information, so any comparison to it is obviously unfair. Nonetheless, it is interesting to include it in the results because its performance is close to the best that can be achieved by any niched EA on this class of problems.

Another noteworthy advantage of MS-NAHC is that it has no parameters, making its application trivial. The same cannot be said of the niched EAs which require a proper setup of the various parameters to work well. In spite of that, even with carefully tuned parameter settings, all but the idealized niching strategy are unable to be faster than multistart hillclimbing.

\begin{figure*}[htbp]
  \center
  \subfigure[$n = 50$]{\includegraphics[width=0.48\linewidth]{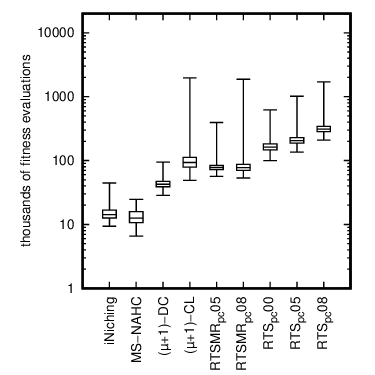}}
  \subfigure[$n = 100$]{\includegraphics[width=0.48\linewidth]{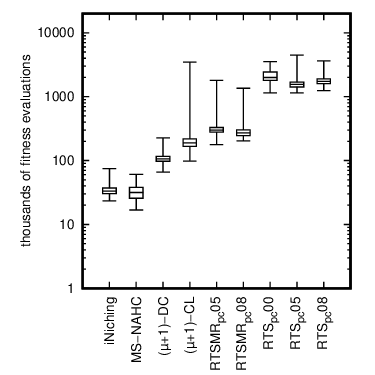}}
  \subfigure[$n = 200$]{\includegraphics[width=0.48\linewidth]{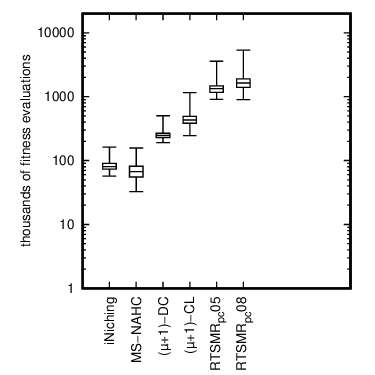}}
  \subfigure[$n = 400$]{\includegraphics[width=0.48\linewidth]{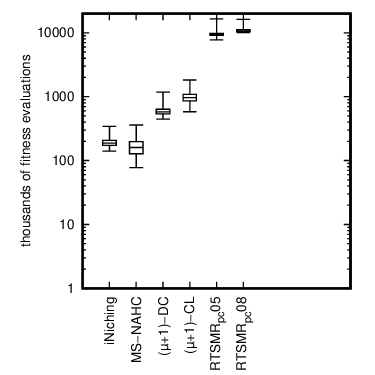}}\\
  \caption{Box-and-whisker plots for the number of fitness evaluations needed to discover all peaks, for instances with 20 peaks and string length $n \in \{50,100,200,400\}$.}
  \label{fig:equal-height-m020}
\end{figure*}

\begin{figure*}[htbp]
  \center
  \subfigure[$n = 50$]{\includegraphics[width=0.48\linewidth]{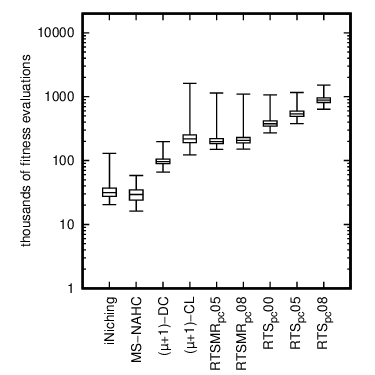}}
  \subfigure[$n = 100$]{\includegraphics[width=0.48\linewidth]{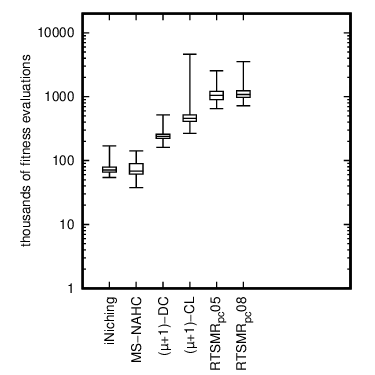}}
  \subfigure[$n = 200$]{\includegraphics[width=0.48\linewidth]{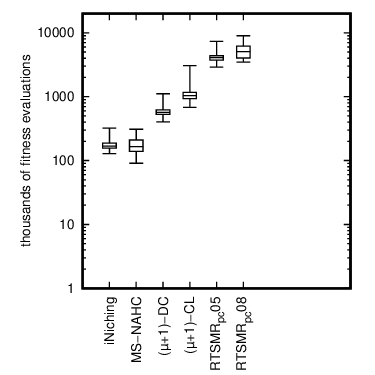}}
  \subfigure[$n = 400$]{\includegraphics[width=0.48\linewidth]{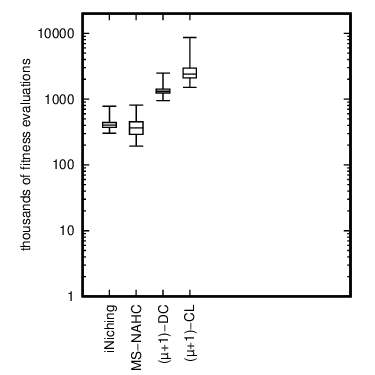}}\\
  \caption{Box-and-whisker plots for the number of fitness evaluations needed to discover all peaks, for instances with 40 peaks and string length $n \in \{50,100,200,400\}$.}
  \label{fig:equal-height-m040}
\end{figure*}

\begin{figure*}[htbp]
  \center
  \subfigure[$n = 50$]{\includegraphics[width=0.48\linewidth]{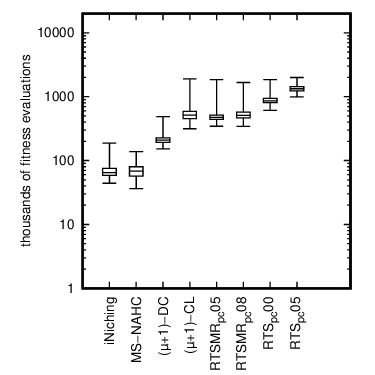}}
  \subfigure[$n = 100$]{\includegraphics[width=0.48\linewidth]{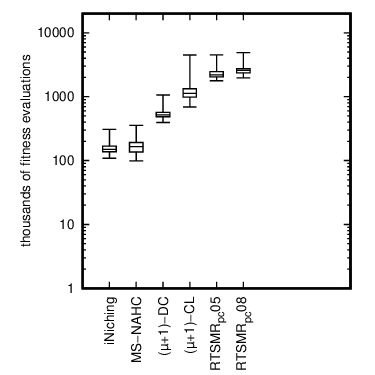}}
  \subfigure[$n = 200$]{\includegraphics[width=0.48\linewidth]{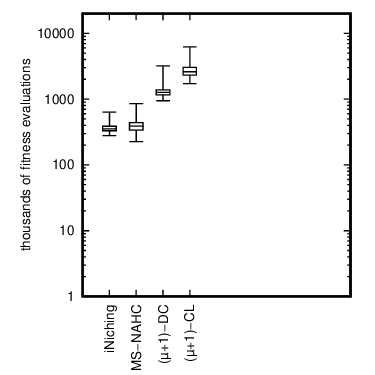}}
  \subfigure[$n = 400$]{\includegraphics[width=0.48\linewidth]{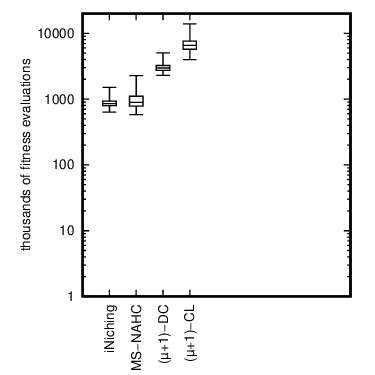}}\\
  \caption{Box-and-whisker plots for the number of fitness evaluations needed to discover all peaks, for instances with 80 peaks and string length $n \in \{50,100,200,400\}$.}
  \label{fig:equal-height-m080}
\end{figure*}

\begin{figure*}[htbp]
  \center
  \subfigure[$n = 50$]{\includegraphics[width=0.48\linewidth]{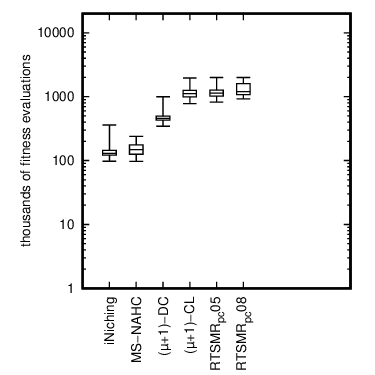}}
  \subfigure[$n = 100$]{\includegraphics[width=0.48\linewidth]{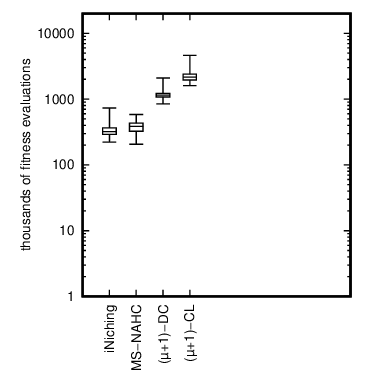}}
  \subfigure[$n = 200$]{\includegraphics[width=0.48\linewidth]{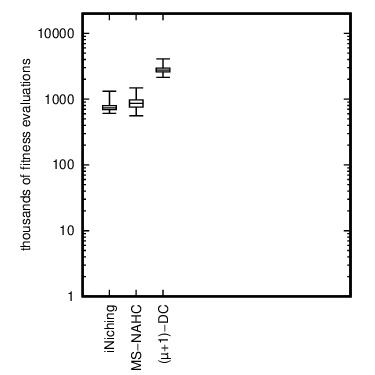}}
  \subfigure[$n = 400$]{\includegraphics[width=0.48\linewidth]{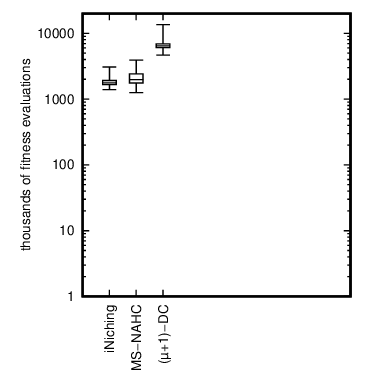}}\\
  \caption{Box-and-whisker plots for the number of fitness evaluations needed to discover all peaks, for instances with 160 peaks and string length $n \in \{50,100,200,400\}$.}
  \label{fig:equal-height-m160}
\end{figure*}

\begin{figure*}[htbp]
  \center
  \subfigure[$n = 50$]{\includegraphics[width=0.48\linewidth]{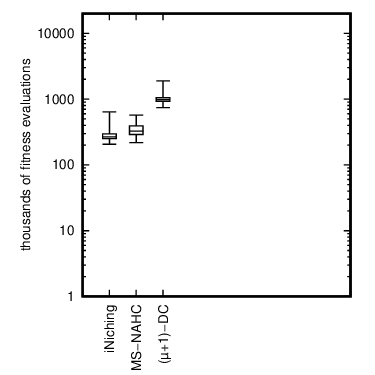}}
  \subfigure[$n = 100$]{\includegraphics[width=0.48\linewidth]{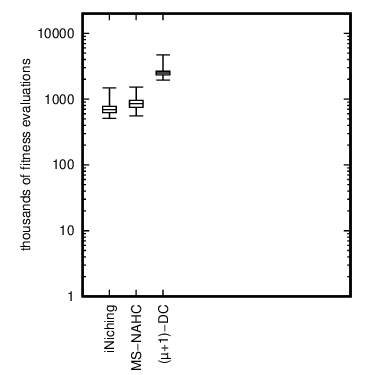}}
  \subfigure[$n = 200$]{\includegraphics[width=0.48\linewidth]{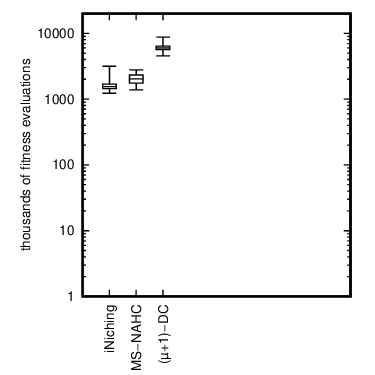}}
  \subfigure[$n = 400$]{\includegraphics[width=0.48\linewidth]{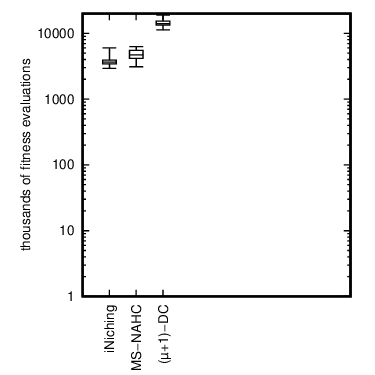}}\\
  \caption{Box-and-whisker plots for the number of fitness evaluations needed to discover all peaks, for instances with 320 peaks and string length $n \in \{50,100,200,400\}$.}
  \label{fig:equal-height-m320}
\end{figure*}


\section{Summary and Conclusions}
\label{sec:conclusions}

This paper presented a runtime analysis of multistart next ascent hillclimbing for solving instances of the multimodal problem generator and confirmed the analysis through computer simulations. It also showed that conventional niching and mating restriction techniques incorporated in an EA are not sufficient to make them competitive with the hillclimbing strategy.

So far, runtime analysis for the time needed to obtain all optima in a multimodal function with EAs have been restricted to bimodal functions~\citep{Osuna2019,Sudholt2020}. The multimodal problem generator provides a nice generalization upon such functions allowing us to study its performance for increasing number of peaks, making it an interesting class of problems for theoretical analysis. We challenge theoreticians to develop such analysis.

We conjecture that practical niched EAs will not be able to be asymptotically faster than a multistart hillclimbing strategy on this class of problems. The reason for this conjecture is due to the lack of structure in the space of local optima on instances of this problem class: when peaks are distributed uniformly at random across the search space, an optimization algorithm cannot exploit information from one optimum to infer where another optimum might be. The structure in this problem class is restricted to the basin of attraction of the peaks, and is as simple as the structure of \textsc{OneMax}. Although there are EAs asymptotically faster than NAHC for climbing a single-peak problem such as \textsc{OneMax} (for example the $(1+(\lambda,\lambda))$ GA~\citep{Doerr2013}), a simple replacement of NAHC with such an EA would not work in a multistart strategy because NAHC knows when it has reached a local optimum (and thus can restart the search from somewhere else) but the EA does not.

The multistart hillclimbing strategy, however, has obvious drawbacks: it can only solve relatively simple problems. For example, it would perform poorly in the massive multimodal problem~\citep{Goldberg:92b} where the global optima are surrounded by millions of local optima, and it would also perform poorly on any problem that requires a solution to jump to another one that is unreachable by the given neighbourhood.

For most real-world and combinatorial optimization problems, there is some sort of structure in the space of local optima. It is this structure that is usually (implicitly) exploited by EAs and other stochastic local search algorithms, and this is the reason why they tend to outperform simpler hillclimbing strategies on such problems.

%
%

There is a general accepted idea that classical optimization methods, such as hillclimbers, are not adequate for locating multiple optimal solutions. Although it is true that starting from different initial solutions does not guarantee a single-point optimization-based algorithm to arrive at different solutions, that by itself does not immediately make such algorithms inadequate for locating multiple optimal solutions. As a matter of fact, this paper shows a particular fitness landscape where a multistart hillclimbing algorithm performs quite well and is unlikely to be outperformed by population-based algorithms. Ultimately, the adequacy of an algorithm to accomplish a given task has to be measured on the grounds of performance, both in theory and practice.

Finally, and in the same spirit of what others have advocated in the past~\citep{Juels:95,Whitley:95}, the results presented in this paper suggest that hillclimbing strategies should not be easily dismissed and should be used as a baseline method when comparing algorithms, even in the case of multimodal optimization.

\section*{Acknowledgements}
This work was supported by Funda\c{c}\~{a}o para a Ci\^{e}ncia e Tecnologia, I.P., Portugal, under project UIDB/04085/2020.

\small

\bibliographystyle{apalike}
\bibliography{refs}

\appendix

\section{Efficient implementation of $(\mu+1)$-CL when $\kappa=1$}
\label{sec:eClearing}

\begin{algorithm}[tbh]
  \SetKwFunction{Mutation}{mutation}
  \SetKwFunction{EfficientClearingProcedure}{eClearing}
  \SetKw{Downto}{downto}
  \DontPrintSemicolon
  $t = 0$\;
  Initialize $P_0$ with $\mu$ individuals chosen u.a.r.\;
  Apply the clearing procedure to $P_0$  (Algorithm~\ref{alg:clearing-procedure})\;
  Sort $P_0$ in decreasing order of fitness.\\
  \While{termination criterion not met}{
    \tcp{Invariant: $P_t$ is sorted in decreasing order of fitness.}
    Choose $x \in P_t$ u.a.r.\;
    $y =$ \Mutation{$x$}\;
    $yp =$ \EfficientClearingProcedure{$P_t, y$}  $\quad$ \tcp{see Algorithm~\ref{alg:efficient-clearing-procedure}}
    Choose $z \in P_t$ with worst fitness u.a.r.\;
    $zp =$ position of $z$ in $P_t$, i.e. $P_t[zp] = z$\;
   \tcp{Check who survives: $y$ or $z$}
   \eIf{$f(y) \ge f(z)$} {    \tcp{$y$ survives, $z$ is removed.}
      \eIf{$f(y) ==  -\infty$}{
           \tcp{both $y$ and $z$ are cleared.}
           $P_t[zp] = y$
      }
      {
           copy $P[yp..zp-1]$ into $P[yp+1..zp]$\;
           $P_t[yp] = y$\;
      }
   }
   {
      \tcp{$z$ survives, $y$ is discarded.}
      \tcp{do nothing.}
   }
  $t = t+1$\;
}
\caption{Efficient implementation of ($\mu$+1)-CL  when $\kappa=1$.}
\label{alg:Efficient-mu1CL}

\end{algorithm}

\begin{algorithm}[tbh]
  \SetKwInOut{Input}{Input}
  \SetKwInOut{Output}{Output}
  \SetKw{Downto}{downto}
  \SetKwInOut{Note}{Note}
  \Input{Population $P[0..\mu-1]$ and individual $y$. Assumes $P$ is sorted in decreasing order of fitness.}
  \Output{Applies the clearing procedure to $P \cup \{y\}$.
     Upon return $P$ remains sorted.
     Return $yp \in [0..\mu]$ such that:
        $f(P[i]) > f(y), \forall i \in [0..yp-1]$ and
        $f(P[i]) \leq f(y), \forall i \in [yp..\mu-1]$.
  }
  \Note{$-\infty$ denotes a constant value lower than the lowest fitness value in the search space.}
  \DontPrintSemicolon
    $yp =$ position $\in [0..\mu]$ such that
        $f(P[i]) > f(y), \forall i \in [0..yp-1]$ and
        $f(P[i]) \leq f(y), \forall i \in [yp..\mu-1]$\;
    \tcp{Check if someone clears $y$.}
    \For{$i = 0$ \KwTo $yp-1$}{
         \If{$D(P[i], y) < \sigma$}{
             $f(y) = -\infty$\;
  	    \textbf{break}
         }
    }
    \If{$f(y) \neq -\infty$}{
       \tcp{$y$ is a winner of a new niche.}
      $clearedIndividuals = \emptyset$\;
       \For{$i = yp$ \KwTo $\mu-1$}{
            \If{$f(P[i]) ==  -\infty$}{
                \textbf{break} \tcp{$f(P[j]) == -\infty, \forall {j>i}$}
            }
            \If{$D(P[i], y) < \sigma$}{
                $f(P[i]) =  -\infty$\;
                add $P[i]$ to $clearedIndividuals$
            }
        }
    \tcp{Sort $P[yp..i-1]$ in linear time.}
    $k = yp$\;
    \For{$j = yp$ \KwTo $i-1$}{
        \If{$f(P[j]) \neq -\infty$}{
            $P[k] = P[j]$\;
            $k = k+1$
         }
    }
    \ForEach{$x \in clearedIndividuals$}{
            $P[k] = x$\;
            $k = k+1$
    }

    }
    \Return $yp$
\caption{eClearing: an efficient implementation of the clearing procedure when $\kappa=1$.}
\label{alg:efficient-clearing-procedure}
\end{algorithm}

\end{document}